\documentclass[acmtog,nonacm]{acmart}

\author{Bonan Liu}
\authornote{Both authors contributed equally to this research.}
\email{bliu404@connect.hkust-gz.edu.cn}
\author{Handi Yin}
\authornotemark[1]
\email{hyin335@connect.hkust-gz.edu.cn}
\affiliation{%
  \institution{HKUST(GZ)}
  \city{Guangzhou}
  \state{Guangdong}
  \country{China}
}
\author{Manuel Kaufmann}
\affiliation{%
  \institution{ETH AI Center, ETH Zürich}
  \city{Zurich}
  \country{Switzerland}}
\author{Jinhao He}
\affiliation{%
  \institution{HKUST(GZ)}
  \city{Guangzhou}
  \state{Guangdong}
  \country{China}
}
\author{Sammy Christen}
\affiliation{%
  \institution{Department of Computer Science, ETH Zürich}
  \city{Zurich}
  \country{Switzerland}}
\author{Jie Song}
\authornote{Corresponding author.}
\affiliation{%
  \institution{HKUST(GZ)}
  \city{Guangzhou}
  \state{Guangdong}
  \country{China}
}
\affiliation{%
  \institution{HKUST}
  \city{Hong Kong}
  \state{Hong Kong}
  \country{China}
}
\author{Pan Hui}
\affiliation{%
  \institution{HKUST(GZ)}
  \city{Guangzhou}
  \state{Guangdong}
  \country{China}
}
\affiliation{%
  \institution{HKUST}
  \city{Hong Kong}
  \state{Hong Kong}
  \country{China}
}

\AtBeginDocument{%
  \providecommand\BibTeX{{%
    \normalfont B\kern-0.5em{\scshape i\kern-0.25em b}\kern-0.8em\TeX}}}

\usepackage{multicol}
\usepackage{multirow}
\usepackage{colortbl}
\begin{document}
\newcommand{\revise}[1]{#1}
\title{EgoHDM: An \revise{Online} Egocentric-Inertial Human Motion Capture, Localization, and Dense Mapping System}

\begin{abstract}
We present EgoHDM, an \revise{online} egocentric-inertial human motion capture (mocap), localization, and dense mapping system. Our system uses 6 inertial measurement units (IMUs) and a commodity head-mounted RGB camera. EgoHDM is the first human mocap system that offers \emph{dense} scene mapping in \emph{\revise{near real-time}}. Further, it is fast and robust to initialize and fully closes the loop between physically plausible map-aware global human motion estimation and mocap-aware 3D scene reconstruction.
To achieve this, we design a tightly coupled mocap-aware dense bundle adjustment and physics-based body pose correction module leveraging a local body-centric elevation map.
The latter introduces a novel terrain-aware contact PD controller, which enables characters to physically contact the given local elevation map thereby reducing human floating or penetration.
We demonstrate the performance of our system on established synthetic and real-world benchmarks.
The results show that our method reduces human localization, camera pose, and mapping accuracy error by 41\%, 71\%, 46\%, respectively, compared to the state of the art.
Our qualitative evaluations on newly captured data further demonstrate that EgoHDM can cover challenging scenarios in non-flat terrain including stepping over stairs and outdoor scenes in the wild. Project page: \href{https://handiyin.github.io/EgoHDM/}{\textcolor{blue}{https://handiyin.github.io/EgoHDM/}}
\end{abstract}

\begin{teaserfigure}
  \includegraphics[width=\textwidth]{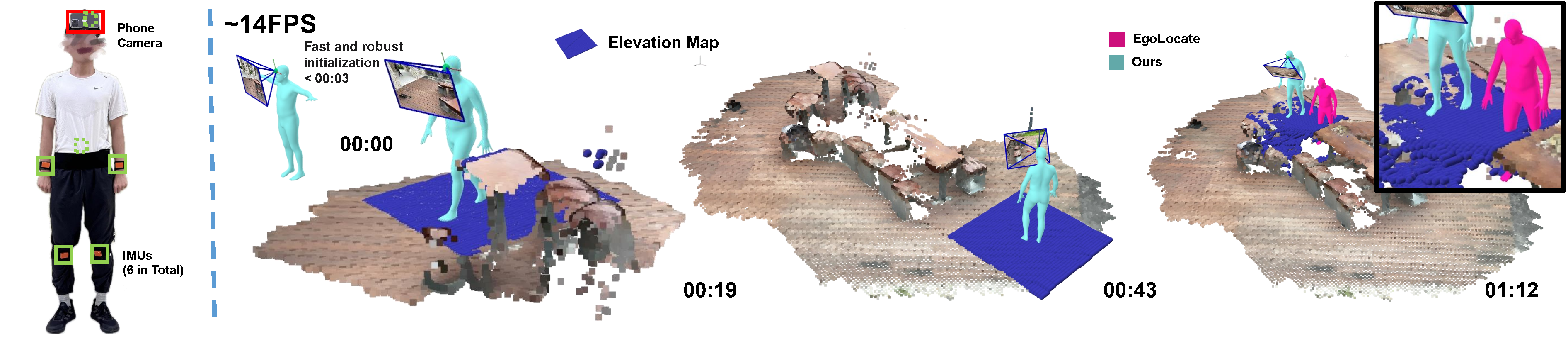}
  \caption{We present an egocentric-inertial human motion capture system that simultaneously estimates a dense map of the scene, runs \revise{in near real-time}, and is fast and robust to initialize. The system takes as input 6 body-worn IMUs and a head-worn RGB camera. It achieves unprecedented accuracy in terms of localization and mapping, and adapts better to non-flat terrain than previous work thanks to physics-based corrections leveraging a local elevation map.}
  \label{fig:teaser}
\end{teaserfigure}

\maketitle

\newcommand{\tableinertial}{
\begin{table}[]
\caption{Comparisons with inertial-based mocap systems TIP \cite{jiang2022transformer}, PIP \cite{yi2022physical} and previous SOTA EgoLocate \cite{EgoLocate2023} on TotalCapture and HPS datasets. The reported numbers are absolute root position errors in meters averaged over all frames.}
\label{arpe}
\resizebox{.45\textwidth}{!}{%
\begin{tabular}{clccccclc}
\toprule
\rowcolor[HTML]{FFFFFF} 
\multicolumn{2}{c}{\cellcolor[HTML]{FFFFFF}}    & \multicolumn{5}{c}{\cellcolor[HTML]{FFFFFF}TotalCapture} &  & HPS  \\ \cmidrule{3-7} \cmidrule{9-9} 
\rowcolor[HTML]{FFFFFF} 
\multicolumn{2}{c}{\multirow{-2}{*}{\cellcolor[HTML]{FFFFFF}Method}}    & acting        & freestyle     & rom       & walking       & average       &  & average       \\
\midrule
\rowcolor[HTML]{EFEFEF} 
\multicolumn{2}{c}{\cellcolor[HTML]{EFEFEF}TIP} & 0.43    & 0.87    & 0.21             & 0.49    & 0.45    &  & 3.00 \\
\rowcolor[HTML]{FFFFFF} 
\multicolumn{2}{c}{\cellcolor[HTML]{FFFFFF}PIP} & 0.61    & 0.51    & \textbf{0.07}    & 0.49    & 0.37    &  & 2.75 \\
\rowcolor[HTML]{EFEFEF} 
\multicolumn{2}{c}{\cellcolor[HTML]{EFEFEF}}    & 0.28    & 0.33    & 0.10             & 0.25    & 0.22    &  & 1.70 \\
\rowcolor[HTML]{EFEFEF} 
\multicolumn{2}{c}{\multirow{-2}{*}{\cellcolor[HTML]{EFEFEF}EgoLocate}} & $\pm$0.06     & $\pm$0.06     & $\pm$0.02 & $\pm$0.03     & $\pm$0.04     &  & $\pm$0.34     \\ \hline
\rowcolor[HTML]{FFFFFF} 
\multicolumn{2}{c}{\cellcolor[HTML]{FFFFFF}Ours}                        & \textbf{0.16} & \textbf{0.18} & 0.09      & \textbf{0.15} & \textbf{0.13} &  & \textbf{1.50} \\ \bottomrule
\end{tabular}%
}
\end{table}
}

% \usepackage{multirow}
% \usepackage{graphicx}
\newcommand{\tablecamera}{
\begin{table}[]
\caption{Comparisons on camera localization results using \cite{ORBSLAM3_TRO} with IMUs (ORB-SLAM3-I) and without (ORB-SLAM3), (on)line and (off)line Droid-SLAM \cite{teed2021droid} and EgoLocate \cite{EgoLocate2023}. The reported numbers are camera localization errors in meters computed over the full sequences. If the SLAM baseline crashes or shows a localization error larger than 20 meters due to fast motions, they are counted as a failure and denoted as ``-''. Note that our method and EgoLocate have no failure cases among all sequences.}
\label{cp}
\resizebox{.45\textwidth}{!}{%
\begin{tabular}{clccccccc}
\toprule
\rowcolor[HTML]{FFFFFF} 
\multicolumn{2}{c}{\cellcolor[HTML]{FFFFFF}}                          & \multicolumn{5}{c}{\cellcolor[HTML]{FFFFFF}TotalCapture} &  & HPS     \\ \cmidrule{3-7} \cmidrule{9-9} 
\rowcolor[HTML]{FFFFFF} 
\multicolumn{2}{c}{\multirow{-2}{*}{\cellcolor[HTML]{FFFFFF}Method}}  & acting     & freestyle  & rom   & walking    & average   &  & average \\ 
\midrule
\rowcolor[HTML]{FFFFFF} 
\multicolumn{2}{c}{\cellcolor[HTML]{FFFFFF}}                          & 0.82       & 0.89       & 0.25  & 0.42       & 0.54      &  & 8.18    \\
\rowcolor[HTML]{FFFFFF} 
\multicolumn{2}{c}{\multirow{-2}{*}{\cellcolor[HTML]{FFFFFF}ORB-SLAM3}} &
  $\pm$0.44 &
  $\pm$0.17 &
  $\pm$0.16 &
  $\pm$0.46 &
  $\pm$0.29 &
   &
  $\pm$1.71 \\
\rowcolor[HTML]{EFEFEF} 
\multicolumn{2}{c}{\cellcolor[HTML]{EFEFEF}}                          & 10.54      & 4.75       & -     & 1.08       & 4.87      &  & -       \\
\rowcolor[HTML]{EFEFEF} 
\multicolumn{2}{c}{\multirow{-2}{*}{\cellcolor[HTML]{EFEFEF}ORB-SLAM3-I}} & $\pm$5.48  & $\pm$2.62  & -     & $\pm$1.88  & $\pm$3.24 &  & -       \\
\rowcolor[HTML]{FFFFFF} 
\multicolumn{2}{c}{\cellcolor[HTML]{FFFFFF} Droid-SLAM (on)}                 & 0.23       & 0.19       & 0.07  & 0.27       & 0.20      &  & -       \\
\rowcolor[HTML]{EFEFEF} 
\multicolumn{2}{c}{\cellcolor[HTML]{EFEFEF}Droid-SLAM (off)}                & 0.14       & 0.10       & 0.07  & 0.24       & 0.14      &  & -       \\
\rowcolor[HTML]{FFFFFF} 
\multicolumn{2}{c}{\cellcolor[HTML]{FFFFFF}}                          & 0.29       & 0.35       & 0.13  & 0.25       & 0.24      &  & 1.69    \\
\rowcolor[HTML]{FFFFFF} 
\multicolumn{2}{c}{\multirow{-2}{*}{\cellcolor[HTML]{FFFFFF}EgoLocate}} &
  $\pm$0.06 &
  $\pm$0.06 &
  $\pm$0.02 &
  $\pm$0.04 &
  $\pm$0.04 &
   &
  $\pm$0.33 \\ \hline
\rowcolor[HTML]{FFFFFF} 
\multicolumn{2}{c}{\cellcolor[HTML]{FFFFFF}Ours} &
  \textbf{0.07} &
  \textbf{0.09} &
  \textbf{0.05} &
  \textbf{0.08} &
  \textbf{0.07} &
  \textbf{} &
  \textbf{1.49} \\ 
\bottomrule
\end{tabular}%
}

\end{table}
}

\newcommand{\tablemapping}{
\begin{table}[]
\caption{Comparison of mapping accuracy with (off)line Droid-SLAM \cite{teed2021droid} and EgoLocate \cite{EgoLocate2023}. The reported numbers are point-to-point distances in meters.}
\centering
\fontsize{6}{6}\selectfont    %{字体尺寸}{行距}
\resizebox{.45\textwidth}{!}{%
\begin{tabular}{cllllll}
\toprule
\multicolumn{2}{c}{}                                                    & \multicolumn{5}{c}{TotalCapture}                                              \\ \cmidrule{3-7} 
\multicolumn{2}{c}{\multirow{-2}{*}{Method}} &
  \multicolumn{1}{c}{acting} &
  \multicolumn{1}{c}{freestyle} &
  \multicolumn{1}{c}{rom} &
  \multicolumn{1}{c}{walking} &
  \multicolumn{1}{c}{average} \\
\midrule
\multicolumn{2}{c}{Droid-SLAM (off)}                                          & 0.73          & 0.72          & 0.51          & 0.84          & 0.72          \\
\rowcolor[HTML]{EFEFEF} 
\multicolumn{2}{c}{\cellcolor[HTML]{EFEFEF}}                            & 0.5           & 0.78          & 0.97          & \textbf{0.41} & 0.66          \\
\rowcolor[HTML]{EFEFEF} 
\multicolumn{2}{c}{\multirow{-2}{*}{\cellcolor[HTML]{EFEFEF}EgoLocate}} & $\pm$0.14          & $\pm$0.30          & $\pm$0.51          & $\pm$0.09          & $\pm$0.25          \\
\midrule
\multicolumn{2}{c}{Ours}                                                & \textbf{0.28} & \textbf{0.35} & \textbf{0.47} & 0.43          & \textbf{0.39} \\ \hline
\end{tabular}%
}
\label{ma}
\end{table}
}

\newcommand{\tableablation}{
\begin{table}[]
\caption{Ablation studies reporting camera localization errors in meters.}
\centering
\fontsize{8.5}{13}\selectfont    %{字体尺寸}{行距}
\resizebox{\columnwidth}{!}{%
\begin{tabular}{ccllcc}
\toprule
                            & \multicolumn{5}{c}{TotalCapture}              \\ \cmidrule{2-6} 
\multirow{-2}{*}{Method}    & acting & freestyle & rom  & walking & average \\ \midrule
\rowcolor[HTML]{EFEFEF} 
Ours w/o SLAM        & 0.61   & 0.50      & 0.07 & 0.48    & 0.37    \\
Ours w/o VIM Initialization & 1.36   & 1.26      & 0.60 & 1.63    & 1.26    \\
\rowcolor[HTML]{EFEFEF} 
Ours w/o Mocap Constraints  & 0.50   & 0.27      & 0.07 & 0.77    & 0.44    \\ \hline
Ours                        & \textbf{0.07}   & \textbf{0.09}      & \textbf{0.05} & \textbf{0.08}    & \textbf{0.07}   \\
\bottomrule
\end{tabular}%
}
\label{ab}
\end{table}
}

\newcommand{\tableinit}{
\begin{table}[]
\caption{\revise{Comparison of initialization time in seconds. The reported number is recorded on our collected data. Note that we exclude T-pose calibration frames for both methods.}}
\centering
\fontsize{8.5}{13}\selectfont
% \resizebox{\columnwidth}{!}{%
\begin{tabular}{cccc}
\toprule
Method    & Classroom  & LHC    & Bench  \\ \hline
EgoLocate & 33.57s & 18.67s & 17.45s \\
Ours      & \textbf{3.06s}  & \textbf{4.41s}  & \textbf{2.88s}  \\ \bottomrule
\end{tabular}%
% }
\label{tab:init}
\end{table}
}
\newcommand{\figurehps}{
\begin{figure*}[ht]
\centering
\includegraphics[width=\textwidth]{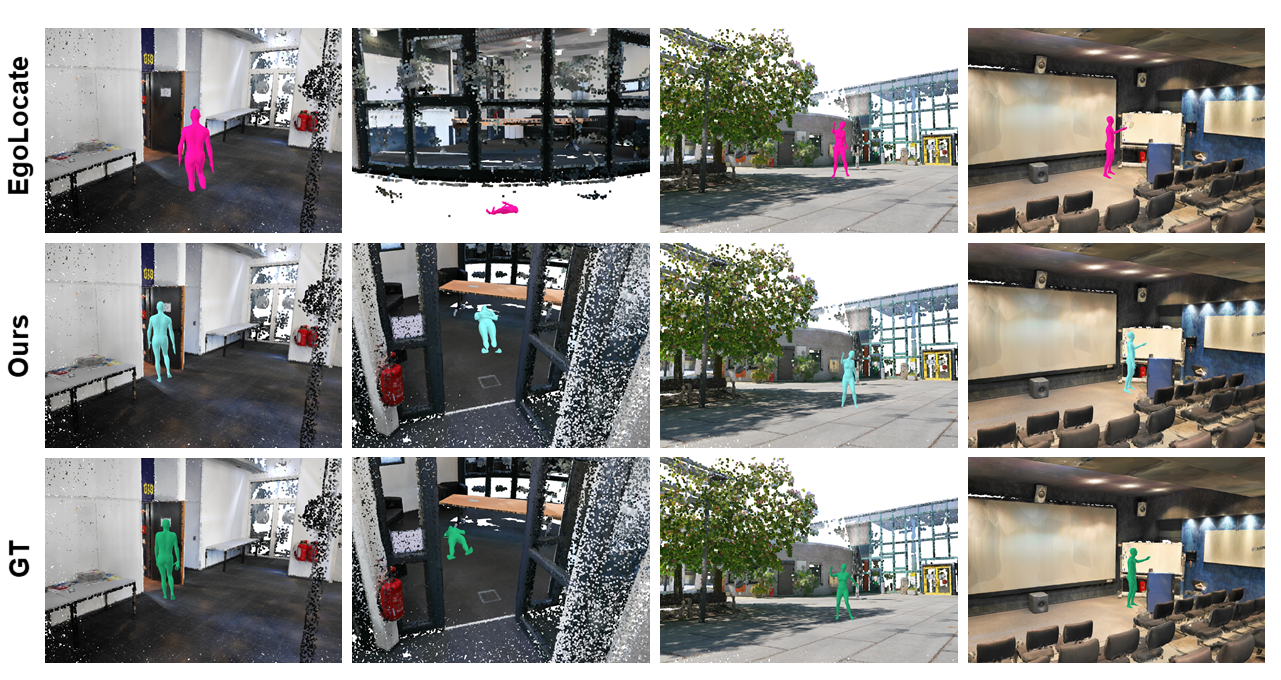}
\caption{Qualitative comparisons on HPS dataset with EgoLocate. We note that EgoLocate estimations can penetrate the floor or float unrealistically, whereas our method estimates more accurate floor contacts, even in the challenging case of the human lying on the floor.}
\label{fig:hps}
\end{figure*}}

\newcommand{\figuretotalcap}{\begin{figure*}[t]
\centering
\includegraphics[width=\textwidth]{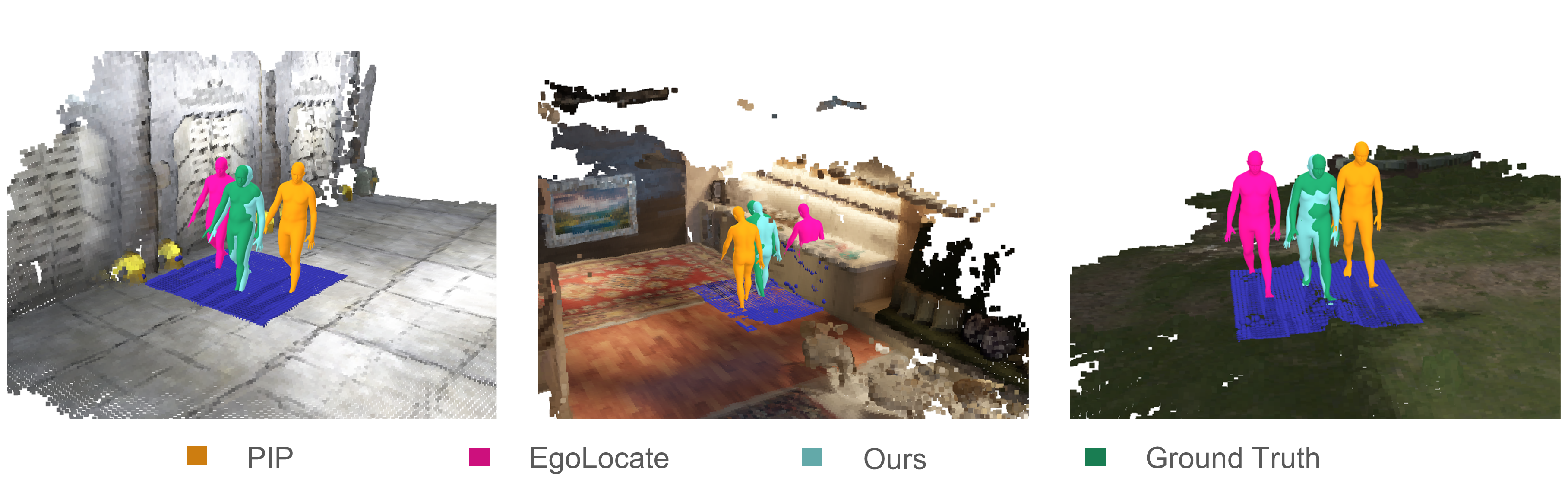}
\caption{Qualitative comparisons on synthetic TotalCapture with PIP (inertial-only) and EgoLocate (inertial + sparse SLAM). The dense map shown in the figure is reconstructed \revise{online} by our system. The blue square represents the elevation map. Our results follow the ground-truth more closely than either baseline.}
\label{fig:totalcap}
\end{figure*}}

\newcommand{\figureunity}{\begin{figure*}[t]
\centering
\includegraphics[width=\textwidth]{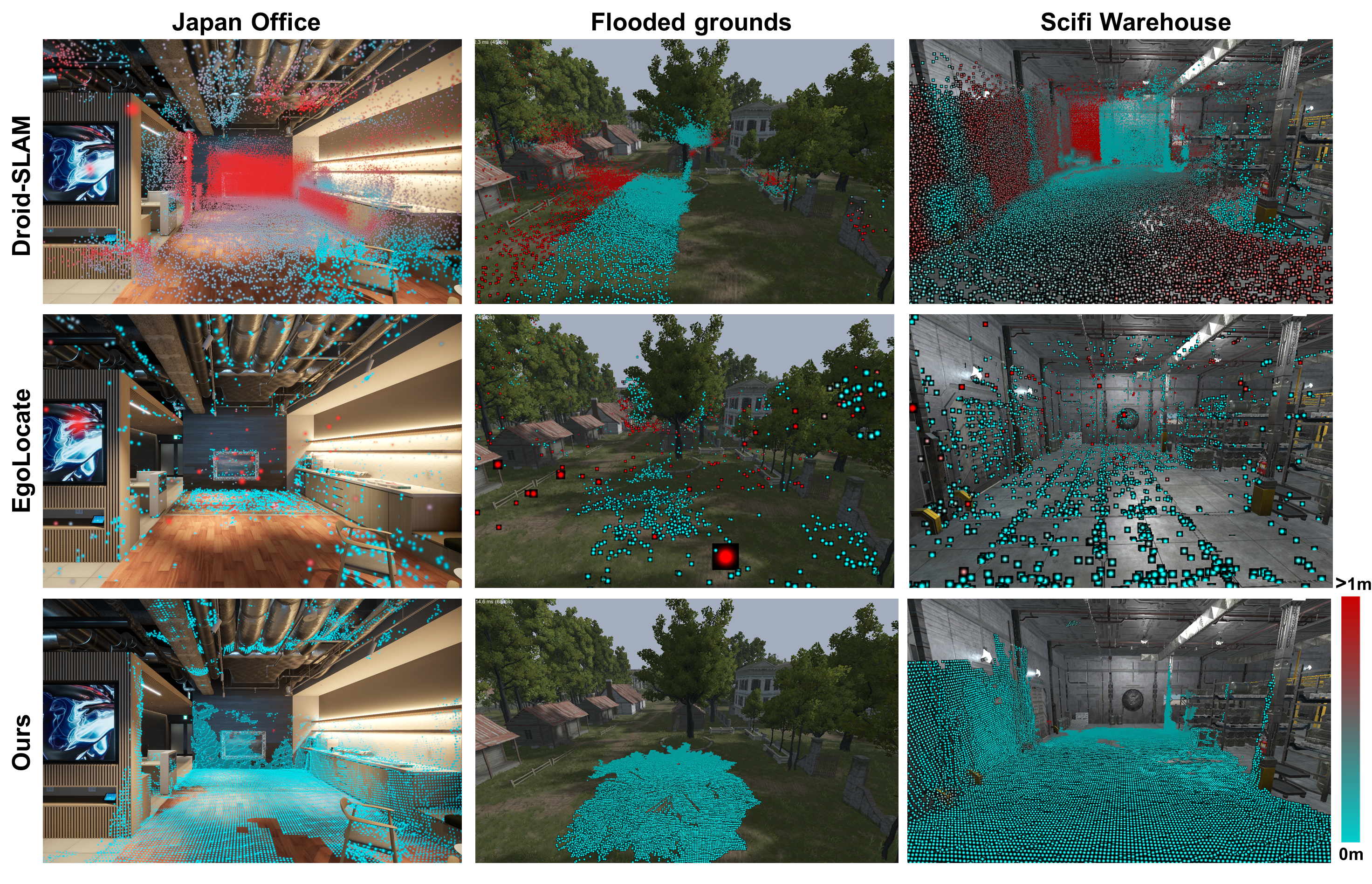}
\caption{Qualitative comparisons of mapping accuracy with offline Droid-SLAM and EgoLocate on synthetic TotalCapture.
For Droid-SLAM, we align the scale with the ground-truth trajectory from the first 8 keyframes.
Blue indicates low, red high error (> 1 meter). Note that even for the challenging ``Flooded Grounds'' scene, our method provides robust mapping of the terrain.}
\label{fig:unity}
\end{figure*}}

\newcommand{\figurelivedemo}{\begin{figure*}[]
\centering
\includegraphics[width=\textwidth]{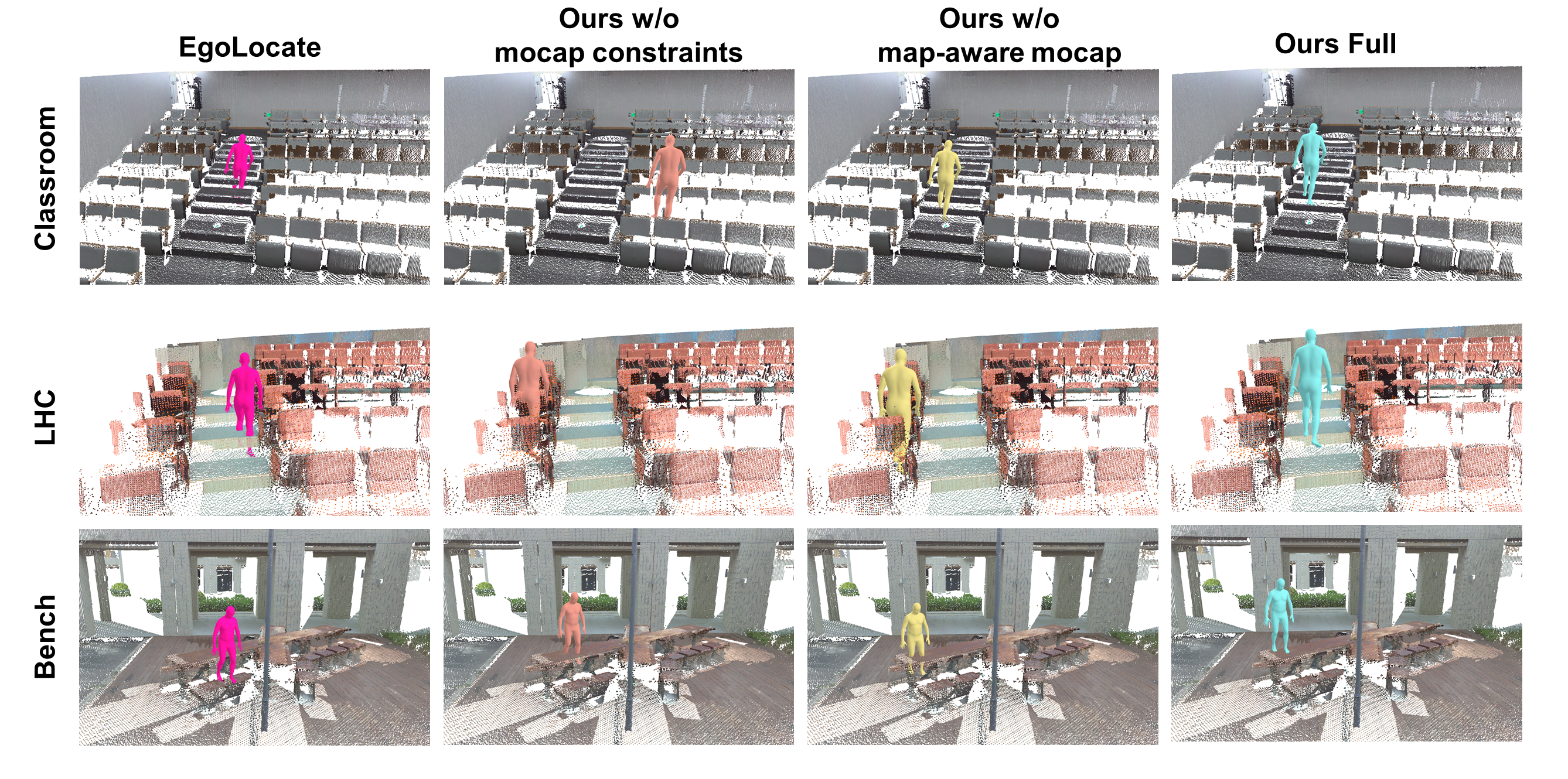}
\caption{Ablation studies on our newly captured sequences involving changing terrain height. Shown are ground-truth LiDAR scans of the scene. We compare EgoLocate (1st column) and a version of our full system that does not use mocap constraints in the MDBA (2nd column) and one that does not use physical correction (3rd column). We notice that in all those baselines unrealistic scene penetrations occur, but not in our full system (4th column).}
\label{fig:ablation}
\end{figure*}}

\newcommand{\figurevim}{\begin{figure}[]
\centering
\includegraphics[width=\columnwidth]{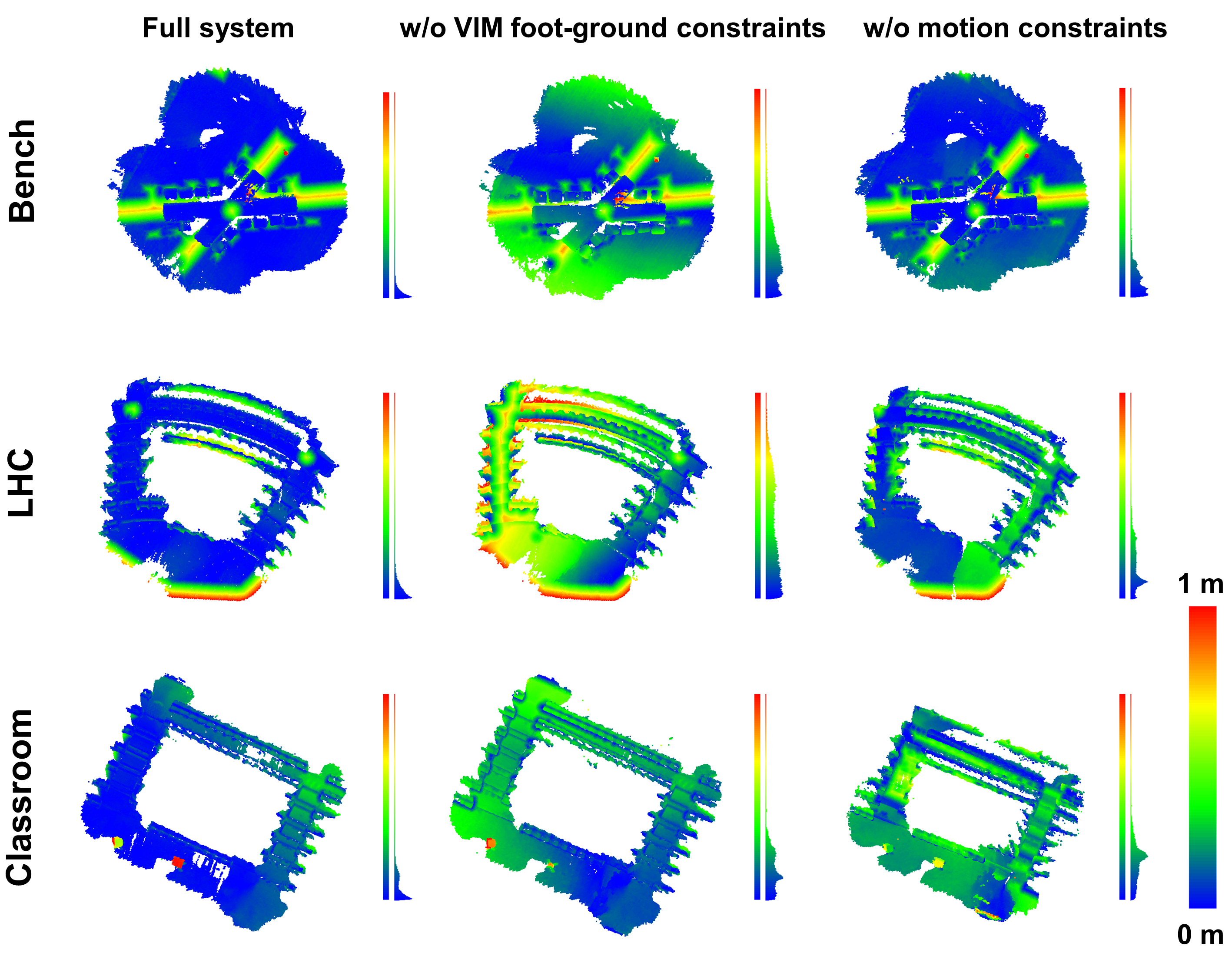}
\caption{\revise{Ablation study in terms of mapping accuracy on our newly captured scenes with terrain height changes.
%We align the ablated reconstructed map with ground-truth LiDAR scans of scene. Blue indicates low, red high error (> 1 meter). 
Errors above $1.0$ m are clipped and excess geometry discarded.
The point-to-point error distribution, drawn next to the color bar, reveals that our full system's error is primarily centered around a low near-zero mean. 
The absence of foot-ground constraints in the VIM initialization (2nd column) and the lack of mocap constraints in the MDBA module (3rd column) lead to increased mapping bias and scale uncertainty, thus driving up the average error and its variance.}}
\label{fig:vim}
\end{figure}}

\section{Introduction}

Striving towards a comprehensive digitization of the real world to enable compelling experiences in mixed reality, it is clear that we have to capture both the human activity \emph{and} the environment.
Unfortunately, human motion capture (mocap) in unconstrained in-the-wild environments is fundamentally challenging in part because existing technology falls short in one or several aspects.
While external camera-based systems may offer high fidelity, especially when deployed in large numbers, they constrain the capture space to a stationary, fixed volume, need careful calibration and struggle with occlusions \cite{chen2020cross,reddy2021tessetrack,shao2022diffustereo,ye2023decoupling,shin2023wham}.
Egocentric capture paradigms, such as the use of body-worn inertial measurement units (IMUs) \cite{luo2021dynamics,yuan2019ego,zhang2021automatic,yi2022physical,jiang2022transformer,huang2018deep}, enable mobile setups and avoid line-of-sight constraints, but they typically suffer from large global drift, making localization in the scene unreliable.
Furthermore, most mocap systems neglect a reconstruction of the environment entirely.
Only recently was it proposed to combine sensor-based mocap with simultaneous scene reconstruction from a head-mounted camera \cite{HPS,EgoLocate2023,lee2024mocap} or with LiDAR sensors \cite{Dai_2022_CVPR}.

The marriage of these two paradigms is interesting because they are conveniently complimentary: the RGB-based localization via SLAM allows for drift-corrected global trajectories and the body-worn sensors deliver body pose that is otherwise difficult to obtain from the forward-facing egocentric camera.
However, combining the two worlds in a way that is mutually beneficial is difficult in practice.
This is because without proper alignment between inertial, body and camera coordinate frames, the \revise{body's} motion constraints might lead to destructive map updates.
While this may be mitigated with physics priors, incorporation of physical constraints necessitates dense mapping systems, which is difficult to come by, especially in online settings.
This is why previous work does not leverage the best of both worlds to the fullest extent: Although scene constraints are used to improve the motion estimation, the motion itself does not inform the scene reconstruction.
Specifically, HPS \cite{HPS} and HSC4D \cite{Dai_2022_CVPR} require a pre-scanned scene and also \cite{lee2024mocap} operate with an offline map that is never updated.
The only work that currently achieves online performance is EgoLocate \cite{EgoLocate2023}. However, also EgoLocate does not fully close the loop because the final pose is not leveraged to update the map.
They also only keep a sparse scene reconstruction and assume a flat ground, leading to body-floor penetrations and poor adaptation to non-flat terrain.

In this paper, we propose the first \revise{\emph{near real-time}} egocentric inertial human localization and mapping system, which simultaneously performs \emph{dense} scene mapping and human motion capture by \emph{jointly optimizing} for the human localization and scene reconstruction and thereby fully closing the loop. Our system, relying on only six IMUs and a head-mounted camera, achieves state-of-the-art performance on several benchmarks both in terms of mapping and human localization error, outperforming both offline and online methods.
Our experiments show that by tightly coupling global motion capture and dense map estimation we can indeed design a system that is mutually beneficial for both tasks.

EgoHDM is enabled by a method that consists of several novel key components.
First, we introduce a new visual-inertial motion (VIM) initialization method to accurately align the inertial and camera coordinate frames. This module explicitly takes body shape into account to better determine scale, and compared to EgoLocate is faster (< 3 seconds) and more robust as it does not require lengthy motion trajectories.
Second, we design a mocap-aware dense bundle adjustment (MDBA) module, which jointly optimizes the camera poses and the depth images of keyframes.
This module tightly couples human motion and body shape priors with RGB-based SLAM.
It further leverages recent advancements in real-time monocular SLAM whereby initialization provided by Droid-SLAM \cite{teed2021droid} is volumetrically fused into dense scene maps weighted by uncertainties provided by probabilistically estimated depth covariance maps \cite{sig}.
Third, we introduce a map-aware physical correction module, which refines poses provided by a learning-based inertial pose estimator \cite{yi2022physical} to satisfy physical foot-to-ground contact constraints.
This is enabled by a 2.5D elevation map, extracted from the dense map in a 2 meter square centered around the human.
This module not only allows the system to handle non-flat terrain well, it also improves the mapping system because the corrected poses are fed back into the MDBA.

Our experiments demonstrate that EgoHDM leads to improvements both compared to visual-only online and offline SLAM systems, as well as its closest related inertial-visual mocap-aware SLAM system, EgoLocate.
Specifically, EgoHDM reduces human localization, camera pose, and mapping errors by 41\%, 71\%, and 46\%. These results suggest that a complete joint modelling of inertial-based global motion estimation and visual-based SLAM is beneficial for both tasks.
In summary, our contributions are:
\begin{itemize}
    \item EgoHDM, an egocentric-inertial human positioning and mapping system using 6 IMUs and a head-mounted camera, which simultaneously estimates global human pose and \emph{dense} 3D scene maps \revise{in near real-time}.
    This is the first method that fully closes the loop between inertial-based global human pose estimation and monocular RGB-based SLAM.
    \item A mocap-aware dense bundle adjustment and a physics-based correction module to establish foot-ground contact on height-varying terrain by means of a local elevation map. 
    \item A novel VIM initialization method, which introduces body shape as an extra scaling constraint to the SLAM system for fast and accurate initialization.
\end{itemize}
\section{Related work}
\subsection{Egocentric Human Pose Estimation}
\paragraph{Camera-Based}
Human pose estimation (HPE) from egocentric cameras divides into works that use downward-facing cameras, either on the chest \cite{jiang2017seeing} or head \cite{luo2021dynamics,yuan2019ego,wang2021estimating,wang2023scene}, or systems leveraging forward-facing cameras \cite{tome2020selfpose,xu2019mo,zhang2021automatic,rhodin2016egocap}.
Downward-looking cameras are advantageous because they capture the full human body in their field of view, sometimes with the help of fish-eye lenses \cite{rhodin2016egocap,xu2019mo}, but this entails frequent self-occlusions and thus reduced accuracy.
Forward-facing setups aim to closely emulate human perception but body parts are frequently out of view, making it difficult to reconstruct arbitrary human motion.
\revise{One line of work uses the egocentric camera as an external view to capture a second person's (and not the wearer's) motion with global translation \cite{liu20214d}, albeit not in real-time.
}
Overall, it is challenging for current egocentric vision approaches to simultaneously estimate human pose and accurate global translation. 

\paragraph*{Sensor-Based}
Another approach for egocentric HPE involves non-visual sensor-based methods that avoid the pitfalls of camera-based methods.
Commercial solutions employ a dense distribution of 17 IMUs to estimate human body pose \cite{Noitom,Xsens}. To increase mobility and reduce setup times, researchers have investigated the use of sparser sensor sets, specifically accelerometer-based \cite{riaz2015motion,slyper2008action,tautges2011motion}, IMU-based 
\cite{von2017sparse,huang2018deep,yi2021transpose,yi2022physical,jiang2022transformer}, and electromagnetic sensor-based methods \cite{kaufmann2021pose,kaufmann2023emdb} have been proposed.
We follow \cite{EgoLocate2023} and use the learning-based component of PIP \cite{yi2022physical} to provide SMPL pose estimates given 6 IMUs.
Others provide human pose from only 6D tracking data of a headset and two hand-held controllers \cite{jiang2022avatarposer,du2023agrol,yang2024divatrack}.
All these methods have partially overcome the pose ambiguity associated with sparse sensors, enabling accurate estimation of local body pose. However, they either offer no or severely drifting global position estimates or do not include scene reconstructions.

\paragraph*{Inertial-Based Sensing with Scene Constraints}
In recent years, there has been a growing interest to combine egocentric camera- and inertial-based methods.
The first work in this direction is HPS \cite{HPS} that utilized 17 IMU sensors to estimate human body pose and employed egocentric RGB image matching for localization of the human in a pre-scanned map.
HSC4D \cite{Dai_2022_CVPR} replaced the RGB camera with LiDAR and successfully reconstructed human motion and the scene simultaneously.
%, exhibiting good consistency.
SLOPER4D \cite{Dai2023Sloper4D} also use LiDAR to reconstruct the scene, but do so from a third-person view.
Very recently \cite{lee2024mocap} proposed a light-weight system that only uses a head-mounted camera and two smartwatches. Like our system, theirs can handle non-flat terrain, but it is not enforced via physically-based losses and the corrected motion does not feed back into the scene map estimation.

Although these works all estimate dense maps, they operate in an offline manner, and sometimes use LiDAR devices increasing instrumentation.
In contrast, EgoLocate \cite{EgoLocate2023} is an online method that uses only 6 IMUs and an egocentric RGB camera, making it our closest related work.
Leveraging sparse ORB-SLAM \cite{ORBSLAM3_TRO} for localization, EgoLocate can estimate drift-reduced human motion by adding relative motion constraints to filter out ill-matching feature points.
This means that EgoLocate only provides a sparse map, which does not allow to model physically-based human-scene interaction leading to poor performance under non-flat terrain.
Furthermore, EgoLocate does not fully exploit the physical interactions between the human body and the environment because camera-corrected human global trajectories are never fed back to update the mocap module.
Our work, EgoHDM, overcomes all of these issues: it performs dense mapping, uses a physically-based correction module to adjust the human motion to non-flat terrain, which is in turn fed back into the system to improve camera estimation. Moreover, we devise a faster and more robust initialization method that considers human body shape and global scale.

\subsection{Visual and Visual-Inertial Dense SLAM}

Simultaneous localization and mapping (SLAM), especially from monocular RGB, is one of the most challenging computer vision problems. The related literature is vast, so we keep discussions to a minimum.
NeRF-based dense SLAM have been shown to deliver accurate performance, but require RGB-D input \cite{zhu2022nice, yang2022vox}, operate at reduced frequencies (5 Hz) \cite{liso2024loopy} or do not include loop closures \cite{rosinol2023nerf}.
\cite{min2021voldor+, teed2021droid} are optical flow-based SLAM systems and achieve impressive trajectory estimations, albeit with an offline BA. \cite{go-slam} extends this to work online and include loop closures.
\cite{sig} employs probabilistic depth uncertainty estimation, derived directly from the information matrix of the BA in Droid-SLAM \cite{teed2021droid} to volumetrically fuse dense depth estimates into the map with reduced noise and in real time.
We adopt the approach of \cite{sig}, enhancing its formulation with additional mocap constraints that trickles down to an updated block camera matrix formulation.
The above systems can still struggle under rapid motion and motion blur. Visual-inertial odometries can address this, e.g.,
\cite{zhang2023bamf, lisus2023towards}.
Nevertheless, these visual-inertial systems require  precise calibration and specific initialization procedures.
In contrast, by incorporating human body shape data, we propose a fast and robust initialization process.

\section{METHOD}
\begin{figure*}[htbp]
\centering
\includegraphics[width=\textwidth]{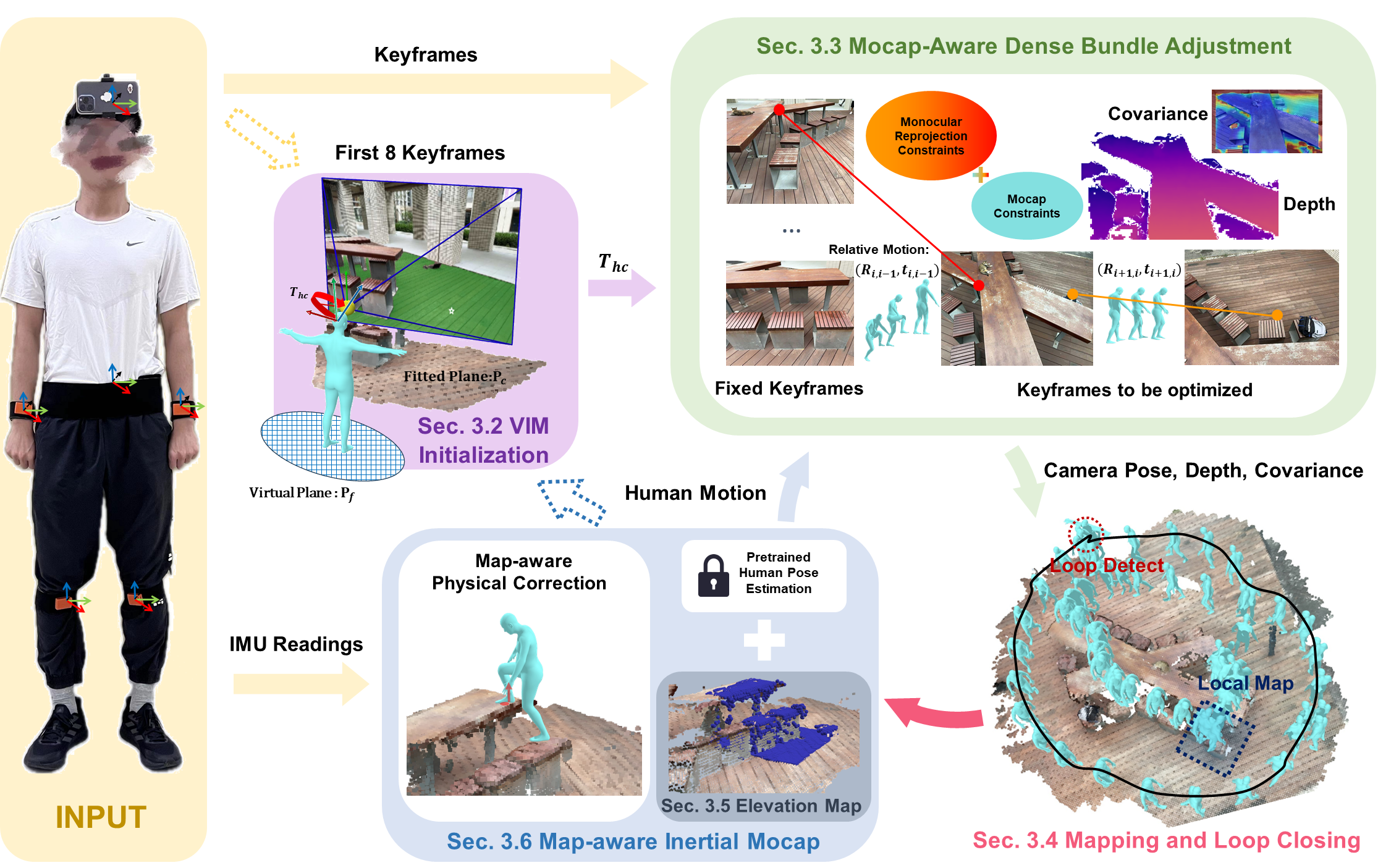}
\caption{\textbf{Overview of EgoHDM.} The inputs to EgoHDM are real-time acceleration and orientation measurements from six body-worn IMUs and monocular egocentric RGB images.
We first initialize the system (VIM Initialization, Sec.~\ref{init}) by finding a similarity transform $\mathbf{T}_{hc}$ that aligns inertial and camera frames with accurate scale found by leveraging body shape constraints.
After initialization, the mocap-aware dense bundle adjustment (MDBA, Sec.~\ref{DBAH}) jointly optimizes camera poses and depth images of keyframes by integrating inertial human motion constraints with RGB-based SLAM \cite{teed2021droid}. \revise{We then construct and maintain a consistent, dense 3D map with global BA and loop closing (Sec.~\ref{mapping}). To reduce the depth noise influence in our global map, covariance-guided volumetric fusion is employed \cite{sig}. Next, we create a local body-centric elevation map with a fixed resolution by projecting the global map along the direction of gravity (Sec.~\ref{localmap}). Lastly, in the map-aware inertial mocap module (Sec.~\ref{mocap}), we refine poses provided by an inertial learning-based pose estimator \cite{yi2022physical} by introducing a physics-based correction module that leverages the elevation map to establish foot-to-ground contact. The corrected poses are fed back to the MDBA, thereby fully closing the loop between inertial-based pose estimation and SLAM-based mapping.}}
\label{fig:method_overview}
\end{figure*}

Our system is an \revise{online} egocentric-inertial human motion capture, localization, and dense mapping framework. It operates by simultaneously reconstructing the environment into a globally consistent dense 3D point cloud map, localizing the human within this map, and generating a body-centric elevation map to model physical foot-ground interactions.
The system's input includes synchronized sensor signals, which consist of inertial data from six IMUs and monocular RGB images from a head-mounted camera.

Our framework seamlessly integrates body-worn inertial-based mocap with a monocular dense mapping system, as illustrated in Fig.~\ref{fig:method_overview}. First, we propose a novel Visual-Inertial Mocap (VIM) initialization that leverages the human body shape as an additional scaling constraint in a short motion sequence for fast and accurate initialization (Sec.~\ref{init}).
Next, we design a mocap-aware dense bundle adjustment (MDBA), which jointly optimizes the camera poses and the depth images of keyframes. This module tightly couples human motion and body shape priors with RGB-based SLAM (Sec.~\ref{DBAH}).
\revise{Then, we discuss loop closing and global bundle adjustment for robust camera pose estimation and long-term map consistency (Sec.~\ref{mapping}). Following this, we introduce a local body-centric elevation map that we extract from the global map (Sec.~\ref{localmap}). We use this elevation map to design a map-aware body pose estimation module that estimates body pose from 6 IMUs and enforces physically correct foot-to-ground contact constraints (Sec.~\ref{mocap})}.

\subsection{Notation and Preliminaries}

Our system takes as input a sequence of 6 IMU measurements $\{([\mathbf{a}_I^1 \dotsc \mathbf{a}_I^6], [\mathbf{R}_I^1 \dotsc \mathbf{R}_I^6])_i\}^{N}_{i=1}$ synchronized with a sequence of egocentric monocular RGB images $\{\mathbf{I}_i\}^{N}_{i=1}$, where $\mathbf{a}_I^k \in \mathbb{R}^3$ denote accelerations, $\mathbf{R}_I^k \in SO(3)$ rotations and $\mathbf{I}_i \in \mathbb{R}^{H_0 \times W_0 \times 3}$. For simplicity we usually only refer to a single sensor and drop superscripts $k$.
From these input measurements, our aim is to estimate the SMPL \cite{loper2023smpl} pose parameters $\boldsymbol{\theta} \in \mathbb{R}^{72}$, translation $\mathbf{t}\in\mathbb{R}^3$ and a dense global map $\mathbf{P}_G\in\mathbb{R}^{N\times3}$ in homogeneous world coordinates.
% \mathbf{Here define $\mathbf{R}^H$ as function of $FK(\mathbf{R}^{root})$}
As our inputs are multi-modal, our approach yields two distinct coordinate frames, \revise{the inertial and the camera coordinate (visual) frame}. In our VIM initialization stage (Sec.~\ref{init}), we calculate the transformation $\mathbf{T}_{hc}$ between these two frames (Fig.~\ref{fig:method_overview}, left). For convenience, after initialization, the world space is set as the camera coordinate frame, and all subsequent notations are adapted to this established world space.
In our mapping system, the camera poses w.r.t. the first input image are denoted as $\{\mathbf{G}_i\}_{i=1}^N$ where $\mathbf{G}_{i} \in SE(3)$. Relative transformations from frame $i$ to $j$ are denoted as $\mathbf{G}_{ij} = \mathbf{G}_j \circ \mathbf{G}_i^{-1}$. The variables $({\mathbf{t}}_{ij}, {\mathbf{R}}_{ij}) \doteq \mathbf{G}_{ij}$ represent the relative position and orientation, respectively. 

\subsection{VIM Initialization with Body Shape Constraint}\label{init}
The goal of the VIM Initialization is to align the coordinate frames involved in our capture setup.
We first employ a T-pose calibration method to compute sensor-to-bone offsets and IMU-to-SMPL frame rotations following \cite{huang2018deep,yi2021transpose,yi2022physical}. For the dense SLAM module, we adopt the keyframe selection and vision-only initialization of Droid-SLAM \cite{teed2021droid}, using the first 8 keyframes, whose indices are stored in $\mathcal{K}$.

Next, we need to find the alignment between the SMPL coordinate frame $\mathcal{F}_\text{SMPL}$, defined as the SMPL root orientation
\revise{
$\mathbf{R}_{r,0}$ and translation $\mathbf{t}_{r,0}$ in the first frame, and the scene coordinate frame $\mathcal{F}_c$, defined as the camera pose $[\mathbf{R}_0 \mid \mathbf{t}_0]$ in the first frame.
}
Because the camera is mounted rigidly on the head, finding this alignment means finding the
\revise{
similarity transformation $\mathbf{T}_{hc} = [s \cdot \mathbf{R}_{hc} \mid \mathbf{t}_{hc}] \in \text{Sim(3)}$
that} maps from the SMPL head joint to the camera. \revise{In other words,} we want to find $\mathbf{T}_{hc}$ that satisfies
\begin{equation}
    \mathbf{G} = \mathbf{T}_{hc} \mathbf{G}_h
\end{equation}
where $\mathbf{G} \in SE(3)$ are camera poses and $\mathbf{G}_h \in SE(3)$ are SMPL head orientation and translation obtained by unrolling the kinematic chain starting from the root, i.e. $\mathbf{G}_h = (\mathbf{R}_h, \mathbf{t}_h) = FK(\mathbf{R}_r, \mathbf{t}_r)$. 

In line with previous research \cite{EgoLocate2023}, we can find $\mathbf{T}_{hc}$ by minimizing $\|\Delta \mathbf{G} \ominus  \Delta (T_{hc} \mathbf{G}_h) \|$ over all keyframes, where $\Delta$ denotes relative transforms w.r.t. the first keyframe and $\ominus$ denotes distance in Sim(3).
Note that we initialize $\mathbf{T}_{hc}$ with an offline estimate that we obtain once before a capture session using an AprilTag \cite{olson2011apriltag1} (more details in supp. mat.).

The above minimization only provides a rough estimation of the scale $s$.
To obtain a more accurate scale estimation, we introduce a novel optimization term that leverages the human body shape and is efficient to compute. 
Assuming the human stands on a flat ground, we construct a virtual plane $\mathbf{P}_f$ located at the base of the SMPL feet that is perpendicular to the upright standing direction. The parameterization of this plane is relative to the head coordinate frame $\mathcal{F}_h$ defined by the SMPL head pose.
At the same time, using the output point cloud from our dense SLAM initialization, we use semantic information to segment the floor area out and then fit a plane $\mathbf{P}_c \in \mathcal{F}_c$ to the masked-out area.
Subsequently, we can find the optimal scale $s$ by minimizing the plane-to-plane distance $d(\cdot)$ between $\mathbf{T}_{hc} \mathbf{P}_f$ and $\mathbf{P}_c$. Hence, overall we minimize 

\begin{equation}
\begin{aligned}
    \mathop{\arg\min}_{\mathbf{T}_{hc}}  \ \
    \alpha \cdot d(\mathbf{P}_c,\mathbf{T}_{hc}\mathbf{P}_f)
    +  \sum_{t \in \mathcal{K}}\beta\cdot \|\Delta \mathbf{G}(t) \ominus  \Delta (\mathbf{T}_{hc} \mathbf{G}_h(t)) \| &&
\end{aligned}
\end{equation}
with $\alpha = 0.9, \beta = 0.1$.
After obtaining the optimal $\mathbf{T}_{hc}$, we define $\mathcal{F}_c$ to be the world space and move all other quantities into it.

\subsection{Mocap-aware Dense Bundle Adjustment} \label{DBAH}
In this section, we discuss our mocap-aware dense bundle adjustment module (MDBA) to tightly couple the depth and camera pose estimation with human mocap constraints.
Our MDBA augments the optical flow-based Droid-SLAM formulation \cite{teed2021droid,teed2020raft} with a novel inertial term $E_{\text{inert}}$.
Specifically, we estimate camera poses and depths by minimizing the following loss function:
\begin{equation}\label{total}
    E_{\text{total}} =  E_{\text{repr}} + \lambda \cdot E_{\text{inert}}
\end{equation}
which weighs the reprojection error $E_{\text{repr}}$ and the inertial error $E_{\text{inert}}$ with weight $\lambda \in \mathbb{R}$.

\noindent\textbf{Reprojection Error.}
Following  \cite{teed2021droid}, we define the reprojection error over the entire frame graph for all image pairs $(i, j) \in \mathcal{E}$.

\begin{equation}
\begin{aligned}
    E_\text{repr} = \ \  &\sum_{(i,j)\in \mathcal{E}} \|\mathbf{u}_{ij}^*-\Pi_c(\mathbf{G}_{ij}\circ\Pi_c^{-1}(\mathbf{u}_i,\mathbf{d}_i)) \|^2_{\Sigma_{ij}},
\end{aligned}
\end{equation}
For each image $\mathbf{I}_i$, the pixel-wise inverse depth is 
defined as $\mathbf{d}_i \in \mathbb{R}^{H_0\times W_0}$.
An image coordinate $\mathbf{u}_i$ with inverse depth $\mathbf{d}_i$ can be reprojected from frame $i$ into frame $j$ according to the warping function $\mathbf{u}^{\prime}_{j}=\Pi_c(\mathbf{G}_{ij}\Pi_c^{-1}(\mathbf{u}_i, \mathbf{d}_i))$, where $\Pi_c$ is the pinhole projection function and $\Pi_c^{-1}$ is its inverse. 
The corresponding points in image $\mathbf{I}_j$ are denoted by $\mathbf{u}_{ij} \in \mathbb{R}^{H_0 \times W_0 \times 2}$. 
Here $\Sigma_{ij} = \text{diag}(\mathbf{w}_{ij})$ represents the confidence weights as predicted following \cite{teed2021droid}.

\noindent\textbf{Inertial Mocap Error.}
In feature-poor environments, during rapid motions, or in case of dynamic obstacles it can be very helpful to employ
a motion prior.  From human inertial mocap (see Sec. \ref{mocap}), we obtain a relative head joint translation prior $\tilde{\mathbf{t}}_{h(i,i-1)}$ and a relative head joint rotation prior $\tilde{\mathbf{R}}_{h(i,i-1)}$. We can transform the relative translation and rotation to the camera frame $\mathcal{F}_c$ to obtain $\tilde{\mathbf{t}}_{i,i-1}$ and $\tilde{\mathbf{R}}_{i,i-1}$. Then $E_{\text{inert}}$ is defined as:

\begin{equation}
\begin{aligned}
    %(\mathbf{u}_i,\mathbf{d}_i)) \|^2_{\sum_{ij}},\\
    E_\text{inert} = &\sum_{(i,i-1)\in \mathcal{E}} \| ({\mathbf{t}}_{i,i-1}-\tilde{\mathbf{t}}_{i,i-1})\|^2_{\Sigma_{\mathbf{t}}}\\
    +&\sum_{(i,i-1)\in \mathcal{E}} \|\log(\tilde{\mathbf{R}}_{i,i-1}^T \mathbf{R}_{i,i-1})^\vee\|^2_{\Sigma_{\mathbf{R}}},
\end{aligned}
\end{equation}
\revise{where $\log(\cdot)^\vee$} maps a rotation matrix to its rotation vector and the covariance $\Sigma_{\mathbf{R}}$, $\Sigma_{\mathbf{t}}$ are set according to the motion prior's uncertainty.

\noindent\textbf{Optimization.}
To solve the constraints defined in Eqn.~\ref{total}, we introduce the Hessian matrix $\mathbf{H}_\text{total}$. Through this matrix, the loss function $E_\text{total}$ can first have a gradient on the keyframe camera pose $\mathbf{G}_i$ and then affect the keyframe inverse depth $\mathbf{d}_i$.
Inspired by \cite{sig}, we utilize the given sparsity pattern of the Hessian to extract a pixel-wise marginal covariance w.r.t. the per-pixel inverse depth \revise{$\Sigma_d$ (see Fig.~\ref{fig:method_overview}, green). This covariance represents the uncertainty of the estimated inverse depth. More details are available in the supp. mat.
Using $\Sigma_d$, we can filter out depths with low confidence, forming the basis for the global map update (Sec.~\ref{mapping}) and the creation of the local elevation map (Sec.~\ref{localmap}).
}

\subsection{Mapping and Loop Closing}
\label{mapping}
\revise{
Given the dense depth images and camera poses computed for each keyframe in the MDBA module (Sec.~\ref{DBAH}), we can now construct a consistent, dense 3D map.
However, the depth images have significant noise due to their high density as depth values are assigned even to textureless areas. 
We thus integrate a well-established volumetric mapping module, proposed by \citet{sig}, into our framework to reduce the depth noise effect on the global map.
We use a hash-based TSDF volumetric representation to fuse the depth maps that we estimated in the MDBA module.
We weigh the SDF values with the depth map associated covariance $\Sigma_d$ and build the global map by sampling from the SDF according to the estimated confidence which allows to maintain map cleanliness.
Additionally, the unavoidable accumulation of camera pose errors might significantly degrade the quality of the map.} Loop closing and global bundle adjustment are thus essential modules for robust pose estimation and long-term map consistency. When a loop is detected, we execute a camera pose-only MDBA - similar to the one in Sec.~\ref{DBAH} but excluding depth optimization - before proceeding to refresh the global map.
Following \cite{go-slam}, we run the MDBA during loop closure in a parallel thread, to ensure efficient loop closing and \revise{online} processing.

\subsection{\revise{Body-Centric Local Elevation Map}}
\label{localmap}

\revise{
Our goal is to estimate body pose that is physically correct and consistent with the current state of the map. 
Even with a dense map, achieving this is non-trivial as the computation should be efficient and the map might have holes. 
To this end, we first introduce a local body-centric elevation map designed for human-scene interactions. This is inspired by \citet{miki2022elevation}, who present a probabilistic elevation map method for robot-centric motion planning.
}

\revise{ More specifically, upon detection of a keyframe, the local map is computed as a 2-by-2 meter uniform grid with a fixed resolution of $M \times M$ cells around the body center (with $M = 100$).
The local elevation map is defined as $\mathbf{P}_L = \{ \mathbf{p}_i \}_{i=1}^{M \cdot M}$, with $\mathbf{p}_i = (x_i, y_i, \hat{h}_i)$. $x_i$ and $y_i$ are the positions obtained when uniformly dividing the grid into $M$ cells along each dimension.
The estimated height $\hat{h}_i$ are the $z$-coordinates obtained from the points of the cropped global map $\mathbf{P}_G$ after projecting them onto each cell $i$ along the direction of gravity specified in the T-pose calibration process (Sec.~\ref{init}). If several points fall into a cell, we take the maximum $z$.
If no points fall into a cell, we interpolate the value using nearest neighbors. This map remains in use until another keyframe is identified, at which point we update the map with the latest data.
}
\subsection{\revise{Map-Aware Inertial Mocap}}
\label{mocap}
\revise{
In this module, we estimate the body pose in a physically correct way by leveraging the local elevation map (Sec.~\ref{localmap}). The module consists of two parts: a learning-based estimation module \revise{to obtain an initial estimate} and a physical correction module.}

\revise{In the first part, we follow the} sparse inertial mocap method PIP \cite{yi2022physical} and utilize their pre-trained weights for learning-based human pose estimation. This component takes 6 IMU accelerations and rotations $\{([\mathbf{a}_I^1 \dotsc \mathbf{a}_I^6], [\mathbf{R}_I^1 \dotsc \mathbf{R}_I^6])_i\}^{N}_{i=1}$ as input and outputs SMPL parameters $\mathbf{q}=[\mathbf{t}, \boldsymbol{\theta}]$ and foot contact probabilities.

Next, in the physical correction module, we are inspired by PIP, which maps the estimated SMPL parameters to rigid body physical models for solving physically plausible motions.
Different from PIP, as our method can reconstruct the \emph{dense} geometry of the scene, we leverage the elevation map and allow our physical correction module to search for human-scene contacts based on global position and foot contact probabilities. To better constrain the estimated contact height $h$ from our elevation map, we introduce a contact PD controller that computes the acceleration component of the gravity direction for the contact joints. 
\begin{equation}
\begin{aligned}
&\dot{\mathbf{r}}_c=\mathbf{J}_c\dot{\mathbf{q}} \\
&\ddot{\mathbf{r}}_{c\downarrow}=k_{p_c}(h-\mathbf{r}_{c\downarrow})-k_{d_c}\dot{\mathbf{r}}_{c\downarrow}
\end{aligned}
\end{equation}

We denote the first-order derivative $\dot{\mathbf{q}}$ as the generalized velocity and $\mathbf{J}_c$ as the contact point Jacobian. $\mathbf{r}_c$ represents the contact point position, while the first-order and second-order derivatives of $\mathbf{r}_c$, i.e., $\dot{\mathbf{r}}_c$ and $\ddot{\mathbf{r}}_c$, represent the corresponding velocity and acceleration and ${\downarrow}$ denotes the component in the direction of gravity. $k_{p_c}$ and $k_{d_c}$ are the corresponding gain coefficients. Typically, these quantities refer to a specific time step $t$, but we omit the time subscript for clarity. When combined with the joint rotation PD controller and joint position PD controller, our physical correction module can effectively produce map-aware motions.
\revise{The corrected poses are fed back to the MDBA to close the loop between inertial-based mocap and SLAM-based mapping.}

\subsection{Implementation Details}
All computations are run on an NVIDIA 4090 with 24GB memory. The core methodology employs PyTorch 2.0 \cite{paszke2019pytorch} along with the Rigid module dynamic library (RBDL) \cite{felis2017rbdl}, while our MDBA is implemented using Pypose \cite{wang2023pypose}.

\section{Experiments}

\subsection{Dataset and Metrics}
\textbf{Datasets.} For quantitative comparison of the global human root translation and camera localization, we follow EgoLocate and evaluate our algorithm on the TotalCapture dataset \cite{trumble2017total} and the HPS dataset \cite{HPS}.
As TotalCapture does not contain egocentric cameras, corresponding data is synthetically generated following \cite{EgoLocate2023}.
We discard HPS sequences with obvious calibration errors and extra-long sequences that exceed 8 minutes.
TotalCapture and HPS barely contain human motion with varying terrain heights.
Hence, we also collected several in-the-wild sequences for qualitative evaluation of non-flat motion trajectories. 
\revise{For this data, we obtain ground-truth maps with a LiDAR scanner, but ground-truth poses are not available as it is in the wild. Thus, this data permits quantitative comparisons in terms of mapping accuracy and qualitative evaluations regarding localization.
}

\noindent\textbf{Evaluation metrics.} We report common metrics to measure EgoHDM's performance.
Specifically, we report the absolute global position error of the human root and the absolute global position error of the camera, averaged over all frames.
We further evaluate mapping accuracy by measuring point-to-point distances between our dense mapping result and the ground-truth scene. 

\noindent\textbf{Baselines}
We compare our results to several state-of-the-art methods in related fields, i.e., TIP \cite{jiang2022transformer} and PIP \cite{yi2022physical} for sparse inertial-only mocap, ORB-SLAM3 \cite{ORBSLAM3_TRO} for monocular  and monocular-inertial sparse SLAM, Droid-SLAM \cite{teed2021droid} for monocular dense SLAM, and EgoLocate \cite{EgoLocate2023} for a real-time inertial mocap and sparse SLAM.
We note that there are currently no open-sourced dense visual-inertial odometry systems, so we are unable to compare our method with dense visual-inertial SLAM algorithms.

\tableinertial
\tablecamera
\tablemapping

\figurehps
\figuretotalcap
\figureunity

\figurevim
\figurelivedemo

\subsection{\revise{Comparisons on Established Benchmarks}}
In this section, we provide quantitative and qualitative comparisons to several baselines on common benchmarks. \revise{Please refer to the supplementary video for more visualizations.}

\subsubsection{Comparison on global mocap results}\label{motion}
We present our quantitative results of absolute root error in Tab.~\ref{arpe}. As demonstrated, our method exceeds SOTA performance, achieving a 41\% improvement and 11\% enhancement on the synthetic TotalCapture and the real-world HPS dataset, respectively.
Please note that due to our learning-based keyframe selection method, our full system is deterministic, which means it will not introduce randomness or performance fluctuation like EgoLocate. 
On the TotalCapture dataset, our performance outperforms all sequences except the ``rom'' motion types. Those sequences are mostly standing with little global movement, which can lead to inaccurate SLAM initialization.
We also provide per-scene absolute root error results for HPS in the supp. mat.
For a visualization of results please refer to Fig.~\ref{fig:hps}, where we show sequences from three different scenes including different subjects with different gender. Compared to EgoLocate, our method significantly reduces body-floor penetrations while achieving on-par or better localization errors.
\revise{We note that TotalCapture actions like ``freestyle'' and ``acting'' comprise challenging motions, such as lying on the floor or jumping, which our system is able to handle well. Please refer to the supp. video for a visualization.}

\subsubsection{Comparison on camera localization.}
To evaluate our camera localization error we compare it with EgoLocate and several SLAM baselines, i.e., ORB-SLAM3 \cite{ORBSLAM3_TRO} (sparse mapper) and its visual-inertial version ORB-SLAM3-I. We also compare to online and offline versions of Droid-SLAM \cite{teed2021droid}.

As demonstrated in Tab.~\ref{cp}, our method outperforms EgoLocate in all scenes by 71\% on synthetic data and achieves 12\% improvement on average on the real-world HPS dataset. 
While traditional SLAM algorithms have decimeter-level error in the TotalCapture dataset, our results only show centimeter-level errors and outperform all previous methods. 
ORB-SLAM3-I seems to have larger errors in both datasets than ORB-SLAM3 without IMUs. This is because all visual-inertial odometry methods have strict restrictions on the initialization stage, and as a result, they may suffer under fast human motion or the hand clap that appears at the start of every HPS sequence for synchronization reasons.
This bad performance of ORB-SLAM3-I confirms from a different angle that our VIM initialization module successfully constrains the scale and extrinsics between the mocap and the dense mapping system.
 
\subsubsection{Comparison on mapping accuracy.}
Tab.~\ref{ma} compares our mapping accuracy with offline Droid-SLAM and EgoLocate on the synthetic TotalCapture dataset.
We follow EgoLocate's evaluation protocol and calculate the average distance between each reconstructed map point and the nearest scene point. Our results demonstrate a 46\% improvement on average compared to EgoLocate, while also outperforming Droid-SLAM in all sequences across all scenes. 

Qualitative results demonstrate an even better improvement, as shown in Fig.~\ref{fig:unity}.
Our method reduces mapping errors in the 3D space and accurately estimates dense map points near the terrain. Notably, even in the highly complex outdoor synthetic scene ``Flooded Grounds'', our method can still provide a robust dense mapping of the terrain. For non-terrain areas, as our method adopts uncertainty filtering, observed far-away objects have no effect on human activities and are filtered out automatically in our algorithm.

\tableablation

\subsection{Ablation Studies}
\subsubsection{\revise{In Terms of Localization Error}}
We perform several ablation studies \revise{w.r.t. camera localization errors on the synthetic TotalCapture dataset, summarized in Tab.~\ref{ab}}.

First we note that estimating camera pose from the inertial head sensor alone (``Ours w/o SLAM'') leads to worse localization. This confirms that our MDBA is indeed helpful, which is not obvious given the ORB-SLAM3-I results in Tab.~\ref{cp}.

Second, \revise{
we evaluate the contribution of our VIM initialization. On row ``Ours w/o VIM initialization'' in Tab.~\ref{ab}, we leave out the VIM initialization and observe that in this case performance drops drastically. This demonstrates that the VIM initialization finds a good alignment between human and inertial frame which is crucial to obtain good overall performance.
Notably, it achieves this all while being significantly faster to compute than corresponding initialization procedures in previous work (see Tab.~\ref{tab:init}).
}

Third, when we leave out mocap constraints (``Ours w/o Mocap constraints''), the system shows a much larger error except for the ``rom'' sequence, which indicates that motion constraints indeed help to estimate the camera pose and corresponding depth. The ``rom'' sequence barely contains any global human motion\revise{, but mostly isolated joint articulations and head movement} and therefore \revise{exploiting mocap constraints has less of an effect}.
Overall, Tab.~\ref{ab} shows that we effectively leverage the best of both worlds: SLAM helps inertial-based localization and mocap helps SLAM-based camera localization - if the two coordinate frames are well aligned.

\subsubsection{\revise{In Terms of Mapping Accuracy}}
\revise{
We also report ablation results in terms of mapping accuracy, for which we use our own dataset as it contains varying terrain heights and thus constitutes the most challenging dataset.
Fig.~\ref{fig:vim} shows the error heatmap and per-point error distribution for 3 in-the-wild scenes. The average error of our full system (1st column) is 5.36 cm.
Fig.~\ref{fig:vim} further shows what happens when we leave out foot-ground constraints in the VIM initialization (2nd column) or motion constraints in the MDBA module (3rd column).
In either case, the average error in mapping accuracy increases to 26.33 and 24.45 cm, respectively.
These results indicate that the lack of foot-ground constraints in the VIM initialization and the absence of mocap constraints in the MDBA module lead to increased mapping bias and scale uncertainty, resulting in significant errors. This further underscores the importance of the careful design of those two modules.
}

\subsection{\revise{Additional Evaluations}}
\revise{In Fig.~\ref{fig:ablation}, we show qualitative comparisons of localization performance on our newly collected in-the-wild dataset with changing terrain heights. Fig.~\ref{fig:ablation} confirms our method's superior performance over state of the art also in this setting. EgoLocate clearly struggles with floor penetrations under changing terrain height (1st column).
We further qualitatively show what happens when we leave out mocap constraints in the MDBA module (2nd column) or when we do not perform physical corrections aided by a body-centric elevation map (3rd column). We conclude that both components are required for accurate localization in non-flat terrain (4th column).}

\revise{Furthermore, we report the time it takes to initalize our system (via the VIM initialization module) on our captured data in Tab.~\ref{tab:init}. It shows that by simply involving human body shape into the VIM initialization module, our system can largely reduce the startup time compared to EgoLocate.}
\tableinit

\subsection{Limitations and Future Work}

\textbf{Pretrained learning-based mocap.}
We borrow the learning-based network from PIP \cite{yi2022physical} with their pretrained weights to initialize local human pose for our physics-based correction. As also reported in WHAM \cite{shin2023wham}, previous learning-based HPE methods tend to soften the motions, e.g., the knees do not fully bend walking up stairs. Although our method can adapt the character to the estimated elevation map surface, our system may still suffer from ``dampened'' local poses.
This issue could result from the current training datasets largely ignoring non-flat environments. 

\noindent\textbf{Quantitative evaluations.} 
For the evaluation of EgoHDM on non-flat terrain, we currently only provide qualitative results, because there are no corresponding datasets with ground-truth poses under meaningfully changing terrain.

\noindent\textbf{Fast motion.} 
Motion blur due to fast human motion is still a significant issue because it will a) reduce mapping and pose accuracy and 
b) result in more keyframes, thus increasing GPU memory usage and loop closure times for long sequences.

\section{Conclusion}
We have presented EgoHDM, a novel egocentric-inertial human motion capture system that simultaneously estimates global human poses and 3D dense scene maps \revise{near real-time} from as little as 6 IMUs and a head-worn commodity RGB camera.
EgoHDM is the first such system that fully closes the loop between inertial-based mocap and monocular visual-based SLAM, demonstrating that the tight coupling of these tasks is mutually beneficial.
Thanks to a novel physics-based correction, EgoHDM estimates motion over non-flat terrain much better than previous work.
We believe egocentric online human localization and dense scene mapping will open exciting new directions in human-scene understanding.

\begin{acks}
This work was supported by the Guangdong Basic and Applied Basic Research Foundation (No.Z2024098), the Guangzhou Municipal Nansha District Science and Technology Bureau (No.2022ZD012) and the Swiss SERI Consolidation Grant 'AI-PERCEIVE'.
The authors thank Skyland Innovation for their extensive help in data collection and hardware design, and Jie Pan and Ya Wen for their support in this project.
Jie Song is the corresponding author.
\end{acks}

\bibliographystyle{ACM-Reference-Format}
\bibliography{Citation}

\end{document}